\documentclass{article}

\usepackage[preprint]{neurips_2026}

\usepackage[utf8]{inputenc} 
\usepackage[T1]{fontenc}    
\usepackage{hyperref}       
\usepackage{url}            
\usepackage{booktabs}       
\usepackage{amsfonts}       
\usepackage{nicefrac}       
\usepackage{microtype}      
\usepackage{xcolor}         
\usepackage{graphicx}
\usepackage{wrapfig}
\usepackage{caption}
\usepackage{sidecap}
\usepackage[ruled,linesnumbered]{algorithm2e}
\usepackage{multirow}
\usepackage{adjustbox}
\newcommand{\rot}[1]{%
  \begin{adjustbox}{angle=45,lap=\width,origin=bl} 
    #1
  \end{adjustbox}%
}
\usepackage{subcaption}
\usepackage{amsmath}
\usepackage{changepage}

\title{Rényi Entropy: A New Token Pruning Metric for Vision Transformers}

\author{%
  Wei-Yuan Su,\qquad Ruijie Zhang$^{\dagger}$,\qquad Zheng Zhang \\
  University of California, Santa Barbara \\
  \texttt{\{wei-yuansu, ruijiezhang\}@ucsb.edu}, \\
  \texttt{zhengzhang@ece.ucsb.edu} \\
}

\begin{document}

\def\thefootnote{$\dagger$}\footnotetext{Project supervision}
\maketitle

\begin{abstract}
    Vision Transformers (ViTs) achieve state-of-the-art performance but suffer from the $O(N^2)$ complexity of self-attention, making inference costly for high-resolution inputs. To address this bottleneck, token pruning has emerged as a critical technique to accelerate inference. Most existing methods rely on the \texttt{[CLS]} token to estimate patch importance. However, we argue that \texttt{[CLS]} token can be unreliable in early layers where semantic representations are still immature. As a result, pruning in the early layer often leads to inaccurate importance estimation and unnecessary information loss. In this work, we propose a training-free token importance metric, namely \textbf{Col-Ln}, which is derived from Rényi entropy that enables the identification of informative tokens from the first layer of the network, thereby enabling more reliable pruning in token reduction. Extensive experiments on ViTs and Large Vision-Language Models (LVLMs) demonstrate that our approach consistently outperforms state-of-the-art pruning methods across diverse benchmarks.
\end{abstract}

\section{Introduction}
\label{sec:intro}

\begin{figure}[t]
  \centering
  \includegraphics[width=1.0\linewidth]{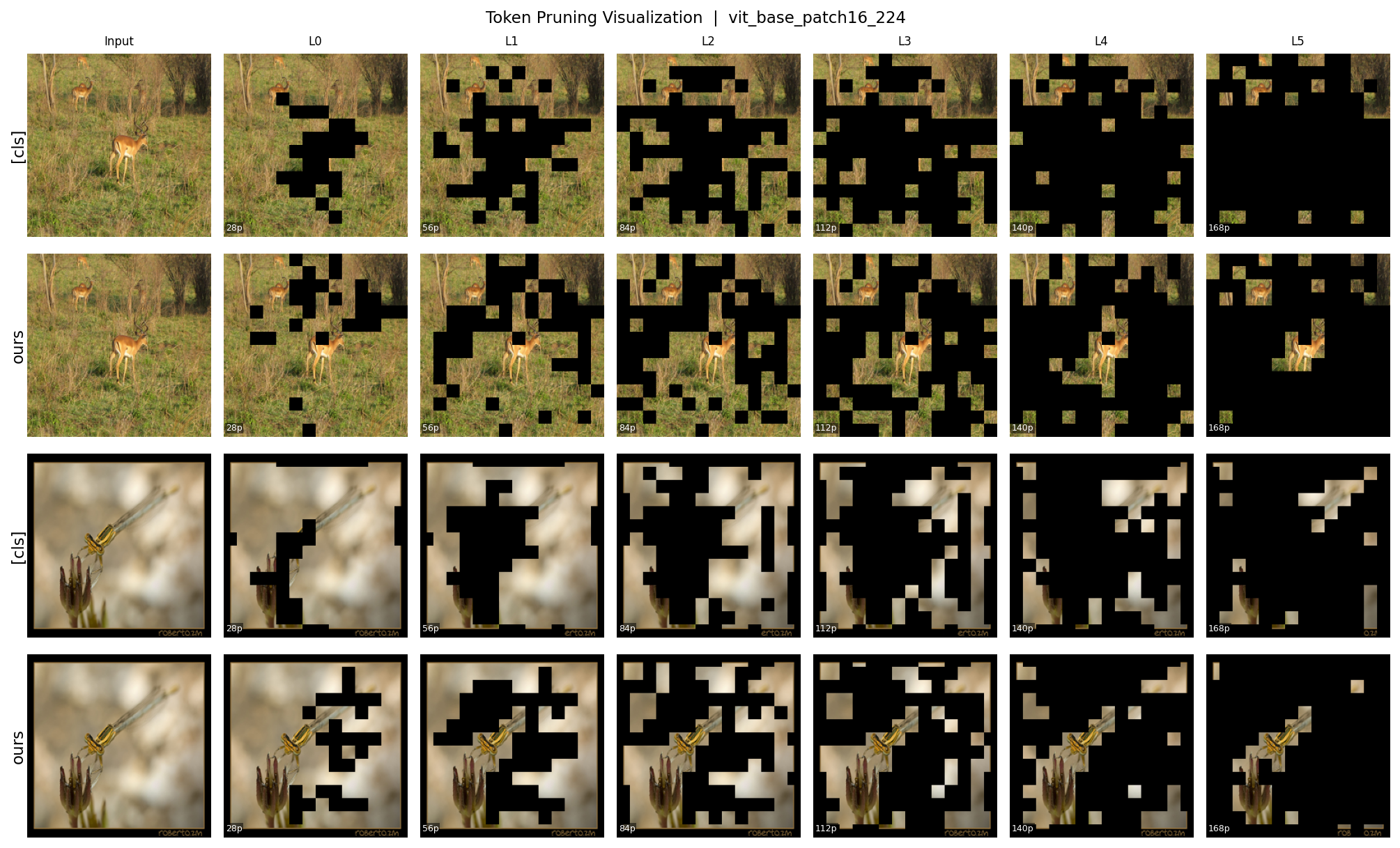}
  \caption{\textbf{Visual Comparison of Token Pruning.} We compare standard \texttt{[CLS]} attention and our proposed metric in the pruning process across initial layers ($L_0$--$L_5$). While \texttt{[CLS]} suffers from semantic immaturity and erroneously prunes critical foreground patches, our method preserves significantly more of the main subject while accurately eliminating redundant background tokens.}
  \label{fig:motivation}
\end{figure}

ViTs \cite{vit} has achieved state-of-the-art performance in computer vision by processing images as sequences of patch tokens via self-attention mechanisms. Although ViTs and LVLMs have demonstrated exceptional capabilities in capturing long-range dependencies, they consume substantial computational costs. Specifically, the computational complexity of the Multi-Head Self-Attention (MHSA) scales quadratically with the number of tokens $N$ [i.e., $\mathcal{O}(N^2)$], making real-time inference prohibitively expensive for applications \cite{attentionisallyouneed,liu2021swin,beltagy2020longformer,zhang2022vsa,zhang2023multi}.

To address this inefficiency, \textit{Token Sparsification} has been investigated. A prominent direction within this domain is \textit{Token Pruning} \cite{evit, tca, prumerge, vispruner, ats, dynamicvit}. A pruning method typically defines an importance score for each token and discards those with the lowest scores. The dominant heuristic, popularized by EViT~\cite{evit}, utilizes the attention weights of the special \texttt{[CLS]} token as a proxy for patch importance. The logical basis is that the \texttt{[CLS]} token interacts with all patches to form a global representation; thus, patches it attends to must be significant.

However, we observe a critical limitation of using the \texttt{[CLS]} token as the pruning metric. In \textbf{early layers} (e.g., layers 0-3), the \texttt{[CLS]} token is initialized randomly and has not yet aggregated sufficient global information. Consequently, its attention map is often misleading or noisy, leading to the misidentification of foreground objects as background (as visualized in Figure~\ref{fig:motivation}).

In this work, we propose \textbf{Column-wise $\ell_n$-norm (Col-Ln)} to address this misidentification issue. \textbf{Col-Ln} departs from the global proxy approach centered on the \texttt{[CLS]} token and instead adopts a collective consensus mechanism driven by patch-to-patch interactions. By leveraging Rényi entropy to characterize attention concentration, \textbf{Col-Ln} effectively quantifies token importance from early layers to the end. As shown in Figure~\ref{fig:motivation}, our proposed metric demonstrates remarkable precision by consistently preserving the core foreground objects.

We summarize our contributions as below:
\begin{enumerate}
    \item We expose the limitations of \texttt{[CLS]}-guided pruning, specifically its instability in early layers.
    
    \item We propose \textbf{Col-Ln}, a parameter-free metric based on Rényi entropy that robustly identifies informative tokens. This enables aggressive pruning from the first layer and effectively addresses the inaccuracy of early-layer importance estimation.
    
    \item We conduct extensive experiments on ViTs and VLMs to evaluate \textbf{Col-Ln} as a pruning metric. In addition, we show that \textbf{Col-Ln} can be integrated into existing pruning methods as a corrective module. Our method consistently surpasses state-of-the-art approaches on all evaluated benchmarks.
\end{enumerate}

\section{The Limitation of \texttt{[CLS]}-based Pruning}
\label{sec:cls_limit}

The standard assumption in ViT pruning is that the \texttt{[CLS]} token serves as a global proxy for image semantics. Therefore, a patch token $x_j$ is considered important if the attention weight $A_{\texttt{[CLS]}, j}$ is high. However, we identify two critical failures in this assumption: 

\noindent
\begin{wraptable}{r}{0.45\textwidth}
    \centering
    \captionof{table}{EViT on ViT-Small Accuracy (\%). Pruning is applied at layers (0, 3, 6).}
    \label{tab:cls_random}
    \small
    \begin{tabular}{c|cc}
        \toprule
        Keep rate ($r$) & \texttt{[CLS]} & Random \\
        \midrule
        0.7 & 66.1 & \textbf{73.2} \\
        0.8 & 74.0 & \textbf{77.6} \\
        0.9 & 79.0 & \textbf{80.0} \\
        \bottomrule
    \end{tabular}
\end{wraptable}
\noindent \enspace$-$ \textbf{Semantic Immaturity in Early Layers:} In the early stages of a ViT, the \texttt{[CLS]} token lacks sufficient semantic grounding as it has not yet aggregated complex spatial dependencies. As visually evidenced in Fig.~\ref{fig:motivation}, our analysis highlights that \texttt{[CLS]}-based metric often targets irrelevant background patches or edge artifacts rather than actual salient objects. This observation is further supported by the quantitative results in Table~\ref{tab:cls_random}, where the \texttt{[CLS]} method fails to even surpass the Random baseline. This catastrophic failure empirically confirms that the global proxy \texttt{[CLS]} at this stage is not just noisy, but potentially misleading by actively discarding critical foreground information. Consequently, existing methods such as EViT \cite{evit} must avoid pruning early layers to maintain stability at the cost of higher FLOPs.

\noindent \enspace$-$ \textbf{Task-specific Attention Bias and Limited Generalizability:} Beyond the immaturity problem, the \texttt{[CLS]} token is inherently supervised by the final task head, making its attention distribution heavily biased toward features relevant only to the training objective. This {task-specific} nature results in poor generalization across diverse or unseen scenarios. As observed in our experiments:

\begin{adjustwidth}{1.5em}{0pt}
\noindent\hangindent=1.5em\hangafter=1 \enspace$\bullet$\enspace \textbf{OOD Sensitivity:} In ImageNet Out-of-Distribution (OOD) tasks, the \texttt{[CLS]} proxy often fails because its importance estimation is over-fitted to specific patterns of the source distribution, leading to the deletion of critical robust features.

\noindent\hangindent=1.5em\hangafter=1 \enspace$\bullet$\enspace \textbf{Cross-dataset Limitations:} In cross-dataset classification, the \texttt{[CLS]} token often inherits the inductive biases of its source pre-training, causing it to focus on coarse global saliency while neglecting the subtle but discriminative features required for accurate domain-specific classification.
\end{adjustwidth}

\noindent This reliance on the pre-training process to achieve reliable pruning limits the applicability of \texttt{[CLS]}-based methods. In contrast, our method captures general importance derived from intrinsic patch-to-patch interactions, ensuring high performance across OOD and cross-dataset benchmarks without requiring task-specific supervision. (Details will be provided in \ref{para:tca})

\section{Methodology}
\label{sec:method}

In this section, we first establish the foundation of our method by leveraging \textbf{Rényi entropy} to quantify token importance through attention concentration. Based on this formulation, we derive a computationally efficient metric, \textbf{Col-Ln}, which serves as a replacement for the \texttt{[CLS]} in token importance estimation. 

\subsection{Rényi Entropy for Importance Measurement}
\label{subsec:theory}

To enable robust token pruning without relying on the \texttt{[CLS]} token, we instead exploit the intrinsic structure of the self-attention mechanism. We assume that informative tokens tend to attract concentrated attention from multiple other tokens, whereas redundant tokens receive more dispersed attention.

We quantify this attention concentration using Rényi entropy\cite{renyi1961measures,zhang2025r}. Let \( \mathbf{A} \in \mathbb{R}^{N \times N} \) denote an attention matrix, where the column \( A_{:,j} \) represents the attention distribution received by token \( j \) from all other tokens. The entropy of this column measures the degree of consensus among tokens regarding the importance of token \( j \).

\begin{itemize}
    \item \textbf{High Entropy (Low Confidence):} The attention received by token $j$ is uniform or scattered. This implies that other tokens are unsure about its importance, or it merely serves as background noise.
    \item \textbf{Low Entropy (High Confidence):} The attention is highly concentrated (spiky). This indicates a strong consensus among other tokens that token $j$ is critical for semantic understanding.
\end{itemize}

We adopt Rényi entropy because it emphasizes high-probability events with a controlled factor $n$, making it well suited for identifying tokens that receive peaked attention. The Rényi entropy of order \( n \) for column \( j \) is defined as

\begin{equation}
    H_n(\text{Column}_j )= \frac{1}{1-n} \log \left( \sum_{i=1}^N (A_{i,j})^n \right)
\end{equation}

The order parameter $n$ regulates the sensitivity of the entropy calculation. Specifically, with a higher $n$, the entropy becomes more sensitive to {\it high-probability} events. This property allows the metric to isolate tokens that have reached a strong collective consensus, while effectively suppressing the background noise that typically characterizes redundant patches.


\begin{SCfigure*}[][t]
	\centering
	\vspace{-15pt}
		\includegraphics[width=0.65\linewidth]{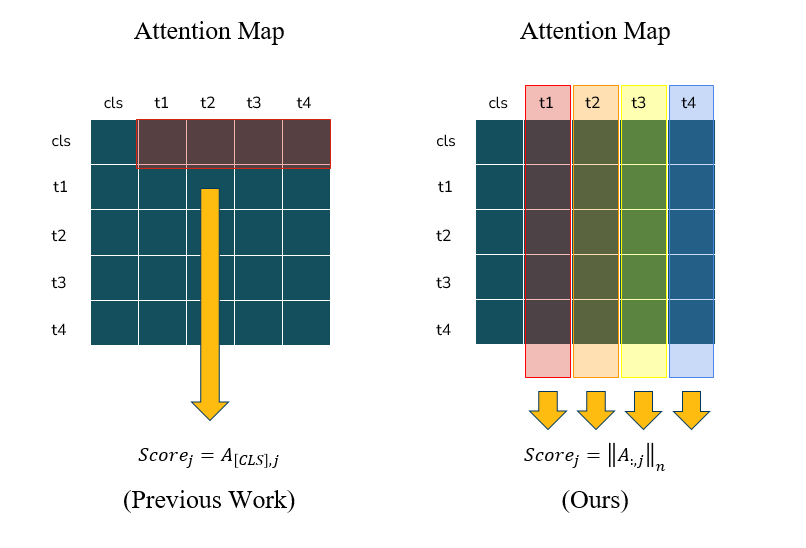}
  \caption{(Left) Conventional methods rely on the \texttt{[CLS]} token as a global proxy, utilizing its attention weights (the top row of the matrix) to estimate patch importance. (Right) Our Col-Ln metric focuses on collective consensus, calculating the column-wise $\ell_n$-norm of an attention matrix.}
  \label{fig:teaser}
\end{SCfigure*}

\subsection{The Col-Ln Metric}
\label{subsec:derivation}

Our objective is to prune these high-entropy (low confidence) tokens and retain those with the lowest entropy. The derivation of our efficient pruning metric, Col-Ln, follows from the mathematical properties of the Rényi Entropy formula:

\begin{equation}
    H_n(\text{Column}_j) = \frac{1}{1-n} \log \left( \sum_{i=1}^N (A_{i,j})^n \right) = \frac{n}{1-n} \log \left( \| A_{:,j} \|_n \right)
\end{equation}
where $\| \cdot \|_n$ denotes the $\ell_n$-norm.

For an order parameter $n > 1$, the coefficient $\frac{n}{1-n}$ is strictly negative. Because the logarithm is a strictly increasing function, the entire term $\frac{n}{1-n} \log(\cdot)$ becomes a strictly decreasing function with respect to its argument. Therefore, identifying tokens with the minimum Rényi entropy is mathematically equivalent to identifying those with the maximum column-wise $\ell_n$-norm ($\| A_{:,j} \|_n$):

\begin{equation}
{\text{Bottom}}-K (H_n(\text{Column}_j), j) \iff \text{Top}-K (\| A_{:,j} \|_n, j)
\end{equation}

This derivation yields our pruning metric, \textbf{Col-Ln} (Column-wise $\ell_n$-norm). By ranking tokens based on the $\ell_n$-norm of their respective attention columns, we efficiently isolate the most informative patches without the heavy computational overhead of explicit entropy calculation. This ensures that the collective consensus is evaluated across all available information channels, allowing the framework to capture the full structural context of the transformer layer. 



\subsection{\texttt{Col-Ln} based Pruning}
\label{subsec:integrated_framework}
As a replacement of CLS-based metric, \textbf{Col-Ln} quantifies the {collective consensus} of all patches to measure the importance of tokens. Following previous analysis, we summarize the algorithm in Algorithm~\ref{alg:col_ln_pruning}.

\begin{algorithm}[t]
\caption{Col-Ln Pruning}
\label{alg:col_ln_pruning}
\KwIn{Input tokens $\mathbf{X} = [\mathbf{x}_{cls}, \mathbf{x}_1, \dots, \mathbf{x}_N] \in \mathbb{R}^{(N+1) \times D}$, keep number $r$, norm degree $n$}
\KwOut{Kept tokens $\mathbf{X}_{out} \in \mathbb{R}^{(r+1) \times D}$}

\BlankLine
$\mathbf{A} = \text{Softmax}\left(\frac{\mathbf{Q}\mathbf{K}^\top}{\sqrt{d_k}}\right) \in \mathbb{R}^{(N+1) \times (N+1)}$\;

\BlankLine
\For{$j = 1$ \KwTo $N$}{
    $S_j = \left( \sum_{i=0}^{N} |A_{i,j}|^n \right)^{\frac{1}{n}}$ 
}
$\mathbf{S} = [S_1, S_2, \dots, S_N] \in \mathbb{R}^{N}$\;

\BlankLine
$\mathcal{I}_{kept} = \text{topk}(\mathbf{S}, k=r)$\;

\BlankLine
$\mathbf{X}_{kept} = \text{Gather}(\mathbf{X}, \mathcal{I}_{kept})$\;

\BlankLine
$\mathbf{X}_{out} = [\mathbf{x}_{cls} , \mathbf{X}_{kept}]$\;

\BlankLine
\Return $\mathbf{X}_{out}$
\end{algorithm}

In addition, we show that \textbf{Col-Ln} can act as a corrective mechanism to preserve critical tokens that may be assigned low importance by the \texttt{[CLS]} token in early layers. The detailed correcting mechanism is summarized in Algorithm~\ref{alg:col_ln_correcting}.

\begin{algorithm}[t]
\caption{Col-Ln Correcting}
\label{alg:col_ln_correcting}
\KwIn{Input tokens $\mathbf{X} = [\mathbf{x}_{cls}, \mathbf{x}_1, \dots, \mathbf{x}_N] \in \mathbb{R}^{(N+1) \times D}$, Keep number $r$, Norm degree $n$, Rescue ratio $c$}
\KwOut{Kept tokens $\mathbf{X}_{out} \in \mathbb{R}^{(r+1) \times D}$}

$k_{cls} = \lfloor r \cdot (1-c) \rfloor$ \;
$k_{col} = r - k_{cls}$ \;

$\mathbf{S}_{cls} = \mathbf{A}_{[0, 1:N]}$ \;
$\mathcal{I}_{cls} = \text{topk}(\mathbf{S}_{cls}, k=k_{cls})$ \;

\For{$j = 1$ \KwTo $N$}{
    $S_j = \left( \sum_{i=0}^{N} |A_{i,j}|^n \right)^{\frac{1}{n}}$ 
}
$\mathbf{S} = [S_1, S_2, \dots, S_N] \in \mathbb{R}^{N}$\;
$\mathbf{S}_{rem} = \{ S_j \mid j \notin \mathcal{I}_{cls} \}$\; 
$\mathcal{I}_{col} = \text{topk}(\mathbf{S}_{rem}, k=k_{col})$ \;

$\mathcal{I}_{kept} = \mathcal{I}_{cls} \cup \mathcal{I}_{col}$ \;
$\mathbf{X}_{kept} = \text{Gather}(\mathbf{X}, \mathcal{I}_{kept})$ \;

$\mathbf{X}_{out} = [\mathbf{x}_{cls} , \mathbf{X}_{kept}]$\;

\Return $\mathbf{X}_{out}$
\end{algorithm}

By retaining such potentially misestimated tokens, the framework mitigates premature information loss and sustains high accuracy even under aggressive pruning budgets.

While our evaluations primarily focus on a \textit{training-free (inference-only)} setting to highlight the metric's intrinsic robustness and plug-and-play capability, \textbf{Col-Ln} is equally well-suited for \textit{fine-tuning}. By avoiding task-specific re-training, we will shown in Section \ref{sec:experiments} that our metric captures fundamental structural properties of attention that generalize across diverse architectures and tasks. Furthermore, when integrated into established fine-tuning frameworks like \textbf{EViT} \cite{evit}, the stable and confirmatory signal provided by collective consensus serves as a superior optimization target, which will be shown in \ref{subsec:evit_finetuning}.




\section{Experiment}
\label{sec:experiments}

We investigate two distinct applications of our metric:
\begin{itemize}
    \item \textbf{Pruning via Col-Ln:} A standalone indicator that directly replaces the standard \texttt{[CLS]} metric to evaluate the intrinsic importance of tokens through collective consensus.
    \item \textbf{Correcting via Col-Ln:} We introduce Col-Ln as a corrective module to refine \texttt{[CLS]}-based pruning approaches. Specifically, it acts as a safeguard against the premature or erroneous token elimination decisions that \texttt{[CLS]}-based methods often make during the initial layers.
\end{itemize}

All our experiments are conducted in a training-free (inference-only) setting. By avoiding task-specific fine-tuning, we demonstrate the robust, intrinsic capability of our metric to identify critical semantic information across diverse architectures and tasks.

\subsection{Experimental Setup}
\label{subsec:setup}

We evaluate the effectiveness of both \textbf{Col-Ln} and the {correcting mechanism} on two representative architectures: standard {Vision Transformers (ViTs)} and {Large Vision-Language Models (LVLMs)}. We conduct image classification on ViT backbones. Efficiency is measured by $GFLOPs$ and Top-1 accuracy. For LVLM, we utilize LLaVA-1.5-7B and evaluate on a comprehensive suite of 10 vision-language benchmarks using LLaVA-1.5-7B \cite{llava, llava-1.5}: VQAv2 \cite{vqav2}, GQA \cite{gqa}, VizWiz \cite{vizwiz}, ScienceQA-IMG (SQA-IMG) \cite{sqa}, TextVQA \cite{textvqa}, POPE \cite{pope}, MME \cite{mme}, MMBench \cite{mmbench}, MMBench-CN, and MM-Vet \cite{mmvet}.

\subsection{Pruning via Col-Ln}
\label{subsec:exp_col_ln}

We first evaluate \textbf{Col-Ln} as a standalone token selection strategy. This setting allows us to observe how effectively collective consensus identifies salient visual features without any task-specific fine-tuning.

\subsubsection{ViT Pruning.}
\label{subsubsec:vit_pruning_col-ln}
We benchmark our Col-Ln metric under two baselines: Naive Pruning and {TCA} \cite{tca}.

\paragraph{Naive ViT Pruning.}
\label{para:naive_pruning_pure_col-ln}

\begin{table}[tb]
\centering
\footnotesize
\begin{minipage}[b]{0.48\textwidth}
\centering
\caption{\textbf{Early-Layer Pruning}}
\label{tab:early_naive_pruning_pure_col-ln}
\setlength{\tabcolsep}{3pt}
\resizebox{\textwidth}{!}{
\begin{tabular}{c|c|cc|cc}
\toprule
\multirow{2}{*}{$Model$} & \multirow{2}{*}{$p$} & \multicolumn{2}{c|}{Accuracy (\%) $\uparrow$} & \multirow{2}{*}{$GFLOPs$} \\
\cmidrule(lr){3-4}
& & \texttt{[CLS]} & \textbf{Ours} & \\
\midrule
\multirow{7}{*}{ViT-S/16} & - & 81.38 & 81.38 & 4.6 \\
& 4 & 80.98 & \textbf{81.14} & 4.2 \\
& 8 & 80.12 & \textbf{80.59} & 3.7 \\
& 12 & 79.06 & \textbf{79.94} & 3.3 \\
& 16 & 77.22 & \textbf{78.78} & 2.9 \\
& 20 & 73.87 & \textbf{76.65} & 2.4 \\
& 24 & 66.54 & \textbf{71.40} & 2.0 \\
\midrule
\multirow{7}{*}{ViT-B/16} & - & 84.54 & 84.54 & 17.6 \\
& 4 & 83.87 & \textbf{84.33} & 15.9 \\
& 8 & 83.07 & \textbf{83.98} & 14.2 \\
& 12 & 82.05 & \textbf{83.47} & 12.6 \\
& 16 & 80.53 & \textbf{82.73} & 11.0 \\
& 20 & 77.93 & \textbf{81.01} & 9.3 \\
& 24 & 72.02 & \textbf{77.29} & 7.7 \\
\midrule
\multirow{4}{*}{ViT-L/16} & - & 85.83 & 85.83 & 61.6 \\
& 8 & 84.82 & \textbf{84.98} & 48.1 \\
& 16 & 82.15 & \textbf{82.94} & 34.8 \\
& 24 & 65.54 & \textbf{72.16} & 21.5 \\
\bottomrule
\end{tabular}}
\end{minipage}
\hfill
\begin{minipage}[b]{0.48\textwidth}
\centering
\caption{\textbf{All-Layer Pruning}}
\label{tab:full_naive_pruning_pure_col-ln}
\setlength{\tabcolsep}{3pt}
\resizebox{\textwidth}{!}{
\begin{tabular}{c|c|cc|c}
\toprule
\multirow{2}{*}{$Model$} & \multirow{2}{*}{$p$} & \multicolumn{2}{c|}{Accuracy (\%) $\uparrow$} & \multirow{2}{*}{$GFLOPs$} \\
\cmidrule(lr){3-4}
& & \texttt{[CLS]} & \textbf{Ours} & \\
\midrule
\multirow{7}{*}{ViT-S/16} & - & 81.38 & 81.38 & 4.6 \\
& 2 & 81.21 & \textbf{81.29} & 4.3 \\
& 4 & 80.92 & \textbf{81.08} & 4.0 \\
& 6 & 80.50 & \textbf{80.76} & 3.7 \\
& 8 & 79.99 & \textbf{80.39} & 3.4 \\
& 10 & 79.22 & \textbf{79.76} & 3.1 \\
& 12 & 78.10 & \textbf{78.63} & 2.9 \\
\midrule
\multirow{7}{*}{ViT-B/16} & - & 84.54 & 84.54 & 17.6 \\
& 2 & 84.19 & \textbf{84.48} & 16.5 \\
& 4 & 83.88 & \textbf{84.30} & 15.3 \\
& 6 & 83.51 & \textbf{84.17} & 14.2 \\
& 8 & 82.98 & \textbf{83.87} & 13.1 \\
& 10 & 82.27 & \textbf{83.39} & 12.0 \\
& 12 & 81.44 & \textbf{82.70} & 10.9 \\
\midrule
\multirow{4}{*}{ViT-L/16} & - & 85.83 & 85.83 & 61.6 \\
& 2 & 85.61 & \textbf{85.68} & 53.8 \\
& 4 & 85.27 & \textbf{85.34} & 46.1 \\
& 6 & 84.40 & \textbf{84.64} & 38.5 \\
\bottomrule
\end{tabular}
}
\end{minipage}
\end{table}

We conduct naive pruning on the ViT backbone (ViT-S, ViT-B, and ViT-L) across various pruning intensities. Specifically, we evaluate our method under two pruning settings: 
(1) \textit{Early-layer pruning}, where tokens are pruned only in the initial six layers to assess its effectiveness in shallow representations; and 
(2) \textit{Standard all-layer pruning}, where tokens are pruned from the first layer through the final layer, following the common practice in prior work. 
This setup allows us to evaluate the effectiveness of \textbf{Col-Ln} both in targeted early pruning and in conventional end-to-end pruning scenarios.

\begin{itemize}
    \item \textbf{Early-Layer Pruning:} In Table~\ref{tab:early_naive_pruning_pure_col-ln}, our Col-Ln metric demonstrates a decisive advantage over the \texttt{[CLS]} baseline across all model scales (ViT-S, ViT-B, and ViT-L). The performance gap becomes increasingly pronounced as the pruning number $p$ increases. At an aggressive budget of $p=24$, our method outperforms the \texttt{[CLS]} baseline by \textbf{+4.86\% on ViT-S, +5.27\% on ViT-B, and a substantial +6.62\% on ViT-L}. These results strongly validate our hypothesis that collective consensus provides a more reliable importance signal in shallow layers than \texttt{[CLS]}.
    \item \textbf{All-Layer Pruning:} To investigate the differences between early-layer pruning and full-layer pruning strategies, we implement all-layer pruning where the final number of kept tokens is maintained to be identical to early-layer pruning. As shown in Table~\ref{tab:full_naive_pruning_pure_col-ln}, our proposed method consistently outperforms the \texttt{[CLS]} baseline across various configurations. Notably, we observe that in certain cases, our method applied specifically to early-layer pruning even surpasses the performance of full-layer pruning at a comparable or even lower computational cost. For instance, our method using early-layer pruning with $p=4$ achieves \textbf{84.33\%} accuracy at \textbf{15.9 GFLOPs}. This performance actually exceeds the \texttt{[CLS]} baseline in the full-layer pruning setting with a more conservative $p=2$, which yields only \textbf{84.19\%} accuracy at a higher computational cost of \textbf{16.5 GFLOPs}.
\end{itemize}

These results demonstrate that our Col-Ln metric is a more faithful indicator of token importance than standard \texttt{[CLS]} attention. More importantly, it overcomes the limitations of previous methods that hindered effective pruning in the early stages.

\paragraph{TCA.} 
\label{para:tca}

To evaluate the cross-domain robustness and generalization of our Col-Ln metric, we integrate the proposed metric into the {TCA} framework \cite{tca}, using {CLIP-ViT-B/16} \cite{clip} as our baseline backbone. We conduct Test-time inference on two challenging benchmarks: {ImageNet Out-of-Distribution (OOD)} sets and a suite of 10 specialized datasets for {cross-dataset classification}.

\begin{itemize}
    \item \textbf{ImageNet OOD Results:} As shown in Table~\ref{tab:tca_ood}, our method consistently enhances the robustness of TCA across all ImageNet \cite{imagenet} variants, including ImageNet-1k \cite{imagenet}, ImageNet-A \cite{imagenet-a}, ImageNet-V2 \cite{imagenet-v2}, ImageNet-R \cite{imagenet-r}, and ImageNet-S \cite{imagenet-s}. Under the (3, 6, 9) pruning configuration at $r=0.7$, our metric improves the average OOD accuracy from \textbf{56.39\%} to \textbf{58.39\%}, a significant \textbf{2.0\%} gain without any fine-tuning. More importantly, when pruning is initiated at earlier layers (0, 3, 6), our method maintains a superior average of \textbf{52.73\%} compared to the baseline's \textbf{51.69\%}.
    
    \item \textbf{Cross-dataset Classification:} We further validate our method on 10 diverse downstream datasets, covering a wide range of different domains: Caltech101 \cite{caltech101}, Oxford Pets \cite{oxford_pets}, Stanford Cars \cite{stanford_cars}, Oxford Flowers \cite{oxford_flowers}, Food101 \cite{food101}, Aircraft \cite{aircraft}, SUN397 \cite{sun397}, DTD \cite{dtd}, EuroSAT \cite{eurosat}, and UCF101 \cite{ucf101}. As summarized in Table~\ref{tab:tca_finegrained}, our approach yields consistent improvements across almost all different domains. Notably, across the entire suite, our metric achieves an average accuracy of \textbf{64.61\%}, surpassing the TCA baseline (\textbf{64.25\%}). In the challenging early-pruning scenario (0, 3, 6), the gap widens to \textbf{1.22\%} (\textbf{61.00\%} vs. \textbf{59.78\%}).
\end{itemize}

\begin{table}[tb]
\centering
\footnotesize
\setlength{\tabcolsep}{10pt}
\caption{Robustness comparison (OOD) using the TCA framework on ImageNet variants ($Keep Rate=0.7$). I, A, V, R, and S denote ImageNet-1K, -A, -V2, -R, and -Sketch, respectively.}
\label{tab:tca_ood}
\begin{adjustbox}{width=\linewidth}
\begin{tabular}{l|c|ccccc|cc}
\toprule
Method & Layers & I & A & V & R & S & \textbf{Average} & $GFLOPs$ \\
\midrule
CLIP-ViT-B/16 & - & 68.37 & 50.20 & 61.86 & 77.55 & 48.24 & 61.24 & 17.6 \\
\midrule
TCA (Baseline) & \multirow{2}{*}{3, 6, 9} & \textbf{65.46} & 47.42 & \textbf{59.14} & 66.04 & 43.91 & 56.39 & \multirow{2}{*}{11.9} \\
\textbf{Ours} & & 65.31 & \textbf{49.27} & 59.13 & \textbf{72.10} & \textbf{45.64} & \textbf{58.39} &  \\
\midrule
TCA (Baseline) & \multirow{2}{*}{0, 3, 6} & \textbf{62.26} & 38.97 & \textbf{55.71} & 60.43 & 41.09 & 51.69 & \multirow{2}{*}{9.2} \\
\textbf{Ours} & & 62.25 & \textbf{40.47} & 55.51 & \textbf{64.33} & 41.09 & \textbf{52.73} &  \\
\bottomrule
\end{tabular}
\end{adjustbox}
\end{table}

\begin{table}[tb]
\centering
\footnotesize
\setlength{\tabcolsep}{5pt}
\caption{Cross-dataset classification performance using TCA ($Keep Rate=0.7$). We report Top-1 accuracy across 10 datasets. Full results for all datasets are provided in the Supplementary Material.}
\label{tab:tca_finegrained}
\begin{adjustbox}{width=\linewidth}
\begin{tabular}{l|c|cccccccccc|c}
\toprule
\rule{0pt}{50pt}
Method & Layers & 
\multicolumn{1}{c}{\rot{Caltech101}} & 
\multicolumn{1}{c}{\rot{Oxford Pets}} & 
\multicolumn{1}{c}{\rot{Stanford Cars}} & 
\multicolumn{1}{c}{\rot{Oxford Flowers}} & 
\multicolumn{1}{c}{\rot{Food101}} & 
\multicolumn{1}{c}{\rot{Aircraft}} & 
\multicolumn{1}{c}{\rot{SUN397}} & 
\multicolumn{1}{c}{\rot{DTD}} & 
\multicolumn{1}{c}{\rot{EuroSAT}} & 
\multicolumn{1}{c}{\rot{UFC101}} & 
\rot{\textbf{Average}} \\
\midrule
CLIP-ViT-B/16 & - & 92.98 & 89.13 & 59.32 & 71.38 & 86.11 & 24.33 & 65.48 & 45.57 & 48.16 & 68.83 & 65.80 \\
\midrule
TCA (Baseline) & \multirow{2}{*}{3, 6, 9} & 90.87 & 85.15 & \textbf{59.32} & \textbf{71.17} & 79.29 & \textbf{23.82} & 62.10 & \textbf{44.92} & 56.89 & 68.97 & 64.25 \\
\textbf{Ours} &  & \textbf{92.74} & \textbf{85.99} & 57.64 & 70.85 & \textbf{81.29} & 22.98 & \textbf{63.14} & 43.97 & \textbf{58.11} & \textbf{69.34} & \textbf{64.61} \\
\midrule
TCA (Baseline) & \multirow{2}{*}{0, 3, 6} & 84.67 & 82.12 & \textbf{52.13} & \textbf{69.59} & 75.95 & 22.05 & 59.02 & 43.03 & 46.93 & 62.33 & 59.78 \\
\textbf{Ours} &  & \textbf{89.90} & \textbf{82.47} & 49.66 & 68.53 & \textbf{77.41} & \textbf{22.17} & \textbf{60.93} & 43.03 & \textbf{48.93} & \textbf{67.01} & \textbf{61.00} \\
\bottomrule
\end{tabular}
\end{adjustbox}
\end{table}

These evaluations demonstrate the robustness and superior generalizability of our metric. Unlike \texttt{[CLS]} attention, which is inherently influenced by task-specific bias and inductive biases from source pre-training, our method captures general importance derived from patch-to-patch interactions. This enables our approach to preserve critical discriminative features across OOD and cross-dataset benchmarks, ensuring high performance without the need for additional fine-tuning.

\subsubsection{LVLM Pruning.}
\label{sec:vlm_exp}

Beyond pure vision tasks, we extend our Col-Ln metric to Large Vision-Language Models (LVLMs) using the LLaVA-1.5-7B model. 

\paragraph{PruMerge.}
\label{para:prumerge}

We evaluate our metric within the {PruMerge} \cite{prumerge} framework on the LLaVA-1.5-7B model. As shown in Table~\ref{tab:vlm_prumerge}, our metric demonstrates a substantial performance advantage over the original PruMerge baseline, particularly in preserving critical semantic information under high compression ratios.

Notably, when the token budget is reduced to $r=256$, our method achieves an average score of \textbf{62.3}, maintaining \textbf{98.3\%} of the full model's performance while significantly outperforming the original PruMerge's 59.2. The gap remains robust even at $r=192$, where our method scores \textbf{61.8}, surpassing the baseline by a remarkable \textbf{+4.0} absolute margin (+6.3\% relative). These results indicate that our metric effectively identifies the essential visual cues that the standard \texttt{[CLS]}-based metric fails to capture in large-scale multimodal contexts.

\begin{table}[tb]
\centering
\caption{Performance comparison of LLaVA-1.5-7B using our metric integrated with the \textbf{PruMerge} framework across representative benchmarks. $r$ denotes the remaining visual tokens.}
\label{tab:vlm_prumerge}
\footnotesize
\setlength{\tabcolsep}{5pt}
\begin{adjustbox}{width=\linewidth}
\begin{tabular}{l|c|cccccccccc|cc}
\toprule
Method & $r$ & VQAv2 & GQA & VizWiz & SQA-IMG & TextVQA & POPE & MME & MMB & MMB-CN & MM-Vet & \textbf{Average} & \textbf{Rel. (\%)} \\
\midrule
LLaVA-1.5 (Full) & 576 & 76.7 & 62.0 & 54.2 & 69.5 & 58.2 & 85.9 & 1505 & 64.6 & 58.1 & 29.7 & 63.4 & 100.0 \\
\midrule
PruMerge & \multirow{2}{*}{256} & 70.8 & 55.7 & \textbf{56.0} & \textbf{68.0} & 54.7 & 73.7 & 1364 & 61.8 & 56.4 & 26.8 & 59.2 & 93.4 \\
\textbf{Ours} &  & \textbf{75.2} & \textbf{59.5} & 54.4 & 67.9 & \textbf{55.4} & \textbf{85.9} & \textbf{1471} & \textbf{63.7} & \textbf{58.5} & \textbf{29.2} & \textbf{62.3} & \textbf{98.3} \\
\midrule
PruMerge & \multirow{2}{*}{192} & 69.0 & 54.3 & \textbf{55.9} & 67.2 & 54.2 & 71.3 & 1298 & 59.3 & 53.0 & 26.6 & 57.8 & 91.2 \\
\textbf{Ours} &  & \textbf{74.6} & \textbf{58.9} & 55.0 & \textbf{67.8} & \textbf{54.7} & \textbf{85.2} & \textbf{1434} & \textbf{63.3} & \textbf{57.0} & \textbf{29.9} & \textbf{61.8} & \textbf{97.5} \\
\bottomrule
\end{tabular}
\end{adjustbox}
\end{table}

\subsection{Correcting \texttt{[CLS]}-pruning via Col-Ln}
\label{subsec:exp_correcting}

This experiment is designed to test whether our metric can successfully rescue critical tokens that the \texttt{[CLS]} token, due to its early-layer semantic immaturity, incorrectly identifies for pruning. 

\subsubsection{ViT Pruning.} 
\label{subsubsec:vit_pruning_correcting}
We implement our Correcting Mechanism under two baselines: {Naive Pruning} baseline and the {EViT} \cite{evit} framework.

\paragraph{Naive ViT Pruning.} 
\label{para:naive_pruning_correcting}

\begin{table}[t]
\centering
\footnotesize
\begin{minipage}[b]{0.48\textwidth}
\centering
\caption{\textbf{Early-Layer Pruning}}
\label{tab:early_naive_pruning_correcting}
\setlength{\tabcolsep}{3pt}
\resizebox{\textwidth}{!}{
\begin{tabular}{c|c|cc|cc}
\toprule
\multirow{2}{*}{$Model$} & \multirow{2}{*}{$p$} & \multicolumn{2}{c|}{Accuracy (\%) $\uparrow$} & \multirow{2}{*}{$GFLOPs$} \\
\cmidrule(lr){3-4}
& & \texttt{[CLS]} & \textbf{Ours} & \\
\midrule
\multirow{7}{*}{ViT-S/16} & - & 81.38 & 81.38 & 4.6 \\
& 4 & 80.98 & \textbf{81.03} & 4.2 \\
& 8 & 80.12 & \textbf{80.65} & 3.7 \\
& 12 & 79.06 & \textbf{79.93} & 3.3 \\
& 16 & 77.22 & \textbf{78.81} & 2.9 \\
& 20 & 73.87 & \textbf{76.53} & 2.4 \\
& 24 & 66.54 & \textbf{71.33} & 2.0 \\
\midrule
\multirow{7}{*}{ViT-B/16} & - & 84.54 & 84.54 & 17.6 \\
& 4 & 83.87 & \textbf{84.34} & 15.9 \\
& 8 & 83.07 & \textbf{84.02} & 14.2 \\
& 12 & 82.05 & \textbf{83.54} & 12.6 \\
& 16 & 80.53 & \textbf{82.68} & 11.0 \\
& 20 & 77.93 & \textbf{81.02} & 9.3 \\
& 24 & 72.02 & \textbf{77.28} & 7.7 \\
\midrule
\multirow{4}{*}{ViT-L/16} & - & 85.83 & 85.83 & 61.6 \\
& 8 & 84.82 & \textbf{84.85} & 48.1 \\
& 16 & 82.15 & \textbf{82.99} & 34.8 \\
& 24 & 65.54 & \textbf{72.69} & 21.5 \\
\bottomrule
\end{tabular}}
\end{minipage}
\hfill
\begin{minipage}[b]{0.48\textwidth}
\centering
\caption{\textbf{Full-Layer Pruning}}
\label{tab:full_naive_pruning_correcting}
\setlength{\tabcolsep}{3pt}
\resizebox{\textwidth}{!}{
\begin{tabular}{c|c|cc|c}
\toprule
\multirow{2}{*}{$Model$} & \multirow{2}{*}{$p$} & \multicolumn{2}{c|}{Accuracy (\%) $\uparrow$} & \multirow{2}{*}{$GFLOPs$} \\
\cmidrule(lr){3-4}
& & \texttt{[CLS]} & \textbf{Ours} & \\
\midrule
\multirow{7}{*}{ViT-S/16} & - & 81.38 & 81.38 & 4.6 \\
& 2 & 81.21 & \textbf{81.32} & 4.3 \\
& 4 & 80.92 & \textbf{81.06} & 4.0 \\
& 6 & 80.50 & \textbf{80.80} & 3.7 \\
& 8 & 79.99 & \textbf{80.46} & 3.4 \\
& 10 & 79.22 & \textbf{79.83} & 3.1 \\
& 12 & 78.10 & \textbf{78.70} & 2.9 \\
\midrule
\multirow{7}{*}{ViT-B/16} & - & 84.54 & 84.54 & 17.6 \\
& 2 & 84.19 & \textbf{84.46} & 16.5 \\
& 4 & 83.88 & \textbf{84.32} & 15.3 \\
& 6 & 83.51 & \textbf{84.15} & 14.2 \\
& 8 & 82.98 & \textbf{83.89} & 13.1 \\
& 10 & 82.27 & \textbf{83.53} & 12.0 \\
& 12 & 81.44 & \textbf{82.82} & 10.9 \\
\midrule
\multirow{4}{*}{ViT-L/16} & - & 85.83 & 85.83 & 61.6 \\
& 2 & 85.61 & \textbf{85.69} & 53.8 \\
& 4 & 85.27 & \textbf{85.28} & 46.1 \\
& 6 & 84.40 & \textbf{84.71} & 38.5 \\
\bottomrule
\end{tabular}
}
\end{minipage}
\end{table}

We evaluate the correcting method on Naive ViT Pruning. From Table~\ref{tab:early_naive_pruning_correcting} and Table~\ref{tab:full_naive_pruning_correcting}, experimental results across ViT-Small, Base, and Large demonstrate that the correcting mechanism effectively rectifies the pruning from the inaccurate token selection of the \texttt{[CLS]} token in early layers. While achieving performance comparable to standalone Col-Ln pruning, this hybrid framework offers enhanced flexibility by allowing a tunable integration of global proxy and collective consensus metrics, effectively leveraging the synergy between these two pruning methods.

\paragraph{EViT.} 
\label{para:evit}

We further integrate our correcting mechanism into the {EViT} framework by replacing its default \texttt{[CLS]}-based importance score with our method. Based on our empirical analysis, we observe two critical phenomena:

\begin{table}[t]
\centering
\footnotesize
\setlength{\tabcolsep}{4pt}
\caption{Comparison of different metrics integrated into EViT on \textbf{ViT-Small}.}
\label{tab:evit_small_results}
\begin{tabular}{c|c|ccc|c}
\toprule
Keep Rate ($r$) & Pruning Layers & \texttt{[CLS]} & \textbf{Ours} & Random & $GFLOPs$ \\
\midrule
ViT-Small  & -- & 81.4 & 81.4 & 81.4 & 4.6 \\
\midrule
\multirow{2}{*}{0.7} & 3, 6, 9 & 78.9 & \textbf{79.0} & 77.3 & 3.0 \\
 & \textbf{0, 3, 6} & 66.1 & \textbf{75.0} & 73.2 & 2.3 \\
 \midrule
\multirow{2}{*}{0.8} & 3, 6, 9 & \textbf{80.5} & \textbf{80.5} & 79.5 & 3.5 \\
 & \textbf{0, 3, 6} & 74.0 & \textbf{78.6} & 77.6 & 2.9 \\
\midrule
\multirow{2}{*}{0.9} & 3, 6, 9 & \textbf{81.3} & \textbf{81.3} & 80.6 & 4.0 \\
 & \textbf{0, 3, 6} & 79.0 & \textbf{80.5} & 80.0 & 3.7 \\
\bottomrule
\end{tabular}
\end{table}

\begin{table}[t]
\centering
\footnotesize
\setlength{\tabcolsep}{4pt}
\caption{Comparison of different metrics integrated into EViT on \textbf{ViT-Base}.}
\label{tab:evit_base_results}
\begin{tabular}{c|c|ccc|c}
\toprule
Keep Rate ($r$) & Pruning Layers & \texttt{[CLS]} & \textbf{Ours} & Random & $GFLOPs$ \\
\midrule
ViT-Base  & -- & 84.5 & 84.5 & 84.5 & 17.6 \\
\midrule
\multirow{2}{*}{0.7} & 3, 6, 9 & \textbf{83.0} & \textbf{83.0} & 81.0 & 11.6 \\
 & \textbf{0, 3, 6} & 69.7 & \textbf{79.9} & 77.8 & 8.8 \\
\midrule
\multirow{2}{*}{0.8} & 3, 6, 9 & 83.9 & \textbf{84.1} & 82.7 & 13.3 \\
 & \textbf{0, 3, 6} & 76.7 & \textbf{82.1} & 81.4 & 11.3 \\
\midrule
\multirow{2}{*}{0.9} & 3, 6, 9 & \textbf{84.5} & \textbf{84.5} & 83.9 & 15.4 \\
 & \textbf{0, 3, 6} & 81.8 & \textbf{83.9} & 83.4 & 14.3 \\
\bottomrule
\end{tabular}
\end{table}

\begin{table}[t]
\centering
\footnotesize
\setlength{\tabcolsep}{4pt}
\caption{The Trend on \textbf{ViT-Small}.}
\label{tab:evit_trend_results}
\begin{tabular}{c|c|ccc|c}
\toprule
Keep Rate ($r$) & Pruning Layers & \texttt{[CLS]} & \textbf{Ours} & Random & $GFLOPs$ \\
\midrule
ViT-Small  & -- & 81.4 & 81.4 & 81.4 & 4.6 \\
\midrule
\multirow{5}{*}{0.7} & 0, 3, 6 & 66.1 & \textbf{75.0} & 73.2 & 2.3 \\
 & 1, 4, 7 & 76.3 & \textbf{77.0} & 74.8 & 2.5 \\
 & 2, 5, 8 & 77.9 & \textbf{78.3} & 75.9 & 2.8 \\
 & 3, 6, 9 & 78.9 & \textbf{79.0} & 77.3 & 3.0 \\
 & 4, 7, 10 & 79.6 & \textbf{79.8} & 78.3 & 3.3 \\
\bottomrule
\end{tabular}
\end{table}

\begin{itemize}
    \item \textbf{The Failure of \texttt{[CLS]} in Early Layers.} A pivotal observation from Table~\ref{tab:evit_small_results} and Table~\ref{tab:evit_base_results} is the catastrophic performance collapse of \texttt{[CLS]}-based pruning when applied to shallow layers (e.g., layers 0, 3, 6). For instance, at a keep rate of $r=0.7$ on ViT-Small, the \texttt{[CLS]} baseline yields a meager \textbf{66.1\%} accuracy, which is significantly lower than the \textbf{73.2\%} achieved by purely random pruning. A similar failure is observed in ViT-Base, where \texttt{[CLS]} (\textbf{69.7\%}) again falls behind the random baseline (\textbf{77.8\%}). However, our proposed metric remains robust across all hierarchies, outperforming the \texttt{[CLS]} baseline by a massive margin of \textbf{+8.9\%} on ViT-Small and \textbf{+10.2\%} on ViT-Base under the most aggressive pruning schedules. These results confirm that our correcting mechanism successfully rescues the framework from the early-stage failures of \texttt{[CLS]}, effectively retrieving informative tokens that the global proxy inaccurately prunes.
    
    \item \textbf{Trend Analysis of pruning layer.} To further investigate the trend of all metrics throughout different pruning strategies, we shift the three-layer pruning schedule from shallow to deep layers while maintaining a constant keep rate of $r=0.7$ on ViT-Small. As summarized in Table~\ref{tab:evit_trend_results}, the results provide clear evidence. While the accuracy of \texttt{[CLS]}-based pruning improves significantly as the pruning schedule shifts toward deeper layers, our proposed metric consistently delivers the best performance regardless of the pruning stage. This shows that our metric remains effective from the very first layer without waiting for \texttt{[CLS]} to aggregate semantic information.
\end{itemize}

\subsubsection{LVLM Pruning.} 
\label{subsubsec:LVLM_pruning_correcting}
We extend our correcting method within the {VisPruner} \cite{vispruner} framework, which focuses on token pruning. 

\paragraph{VisPruner}
\label{para:vispruner}

\begin{table}[t]
\centering
\caption{Performance comparison of LLaVA-1.5-7B using our metric integrated with the {VisPruner} framework across representative benchmarks. $r$ denotes the remaining visual tokens.}
\label{tab:vlm_vispruner}
\footnotesize
\setlength{\tabcolsep}{5pt}
\begin{adjustbox}{width=\linewidth}
\begin{tabular}{l|c|cccccccccc|cc}
\toprule
Method & $r$ & VQAv2 & GQA & VizWiz & SQA-IMG & TextVQA & POPE & MME & MMB & MMB-CN & MM-Vet & \textbf{Average} & \textbf{Rel. (\%)} \\
\midrule
LLaVA-1.5 (Full) & 576 & 76.7 & 62.0 & 54.2 & 69.5 & 58.2 & 85.9 & 1505 & 64.6 & 58.1 & 29.7 & 63.4 & 100.0 \\
\midrule
VisPruner & \multirow{2}{*}{256} & 76.0 & 60.4 & 53.9 & 68.4 & 57.7 & 86.4 & 1454 & 63.4 & 57.2 & 31.1 & 62.7 & 98.9 \\
\textbf{Ours} &  & 76.0 & 60.4 & \textbf{54.0} & \textbf{69.1} & \textbf{57.9} & \textbf{86.5} & \textbf{1483} & 63.4 & \textbf{58.0} & \textbf{32.1} & \textbf{63.2} & \textbf{99.7} \\
\midrule
VisPruner & \multirow{2}{*}{192} & 75.2 & 59.4 & 54.6 & 68.7 & 57.4 & 85.7 & 1459 & 62.3 & 57.2 & 30.5 & 62.4 & 98.4 \\
\textbf{Ours} &  & 75.2 & 59.4 & 54.6 & 68.7 & \textbf{57.8} & \textbf{86.0} & \textbf{1482} & \textbf{63.1} & \textbf{57.9} & \textbf{33.3} & \textbf{63.0} & \textbf{99.4} \\
\bottomrule
\end{tabular}
\end{adjustbox}
\end{table}

As shown in Table~\ref{tab:vlm_vispruner}, our proposed metric consistently enhances pruning accuracy compared to the native importance score across various token budgets. At a token budget of $r=256$, our method achieves an average score of \textbf{63.2}, maintaining \textbf{99.7\%} of the original LLaVA performance, whereas the baseline VisPruner achieves 62.7 (98.9\%). When the budget is further reduced to $r=192$, the performance gap remains, with our metric reaching \textbf{63.0} compared to the baseline's 62.4. These results confirm that our metric provides a more reliable estimation of token significance even in VLM pruning pipelines, ensuring that critical visual information is preserved during compression.

\subsection{EViT Fine-tuning Result}
\label{subsec:evit_finetuning}

To further evaluate the effectiveness of our proposed metric in training-based scenarios, we integrate our metric into the {EViT} framework using the {ViT-Small} backbone. We perform fine-tuning for 30 epochs on ImageNet-1K. For all layers involving token reduction, we maintain a consistent keep rate of $0.7$ across different methods. We compare our metric against the native \texttt{[CLS]} attention metric and a random pruning baseline under two different pruning schedules: $(0, 3, 6)$ and $(3, 6, 9)$.

\begin{figure}[t]
    \centering
    \begin{subfigure}{0.48\textwidth}
        \includegraphics[width=\linewidth]{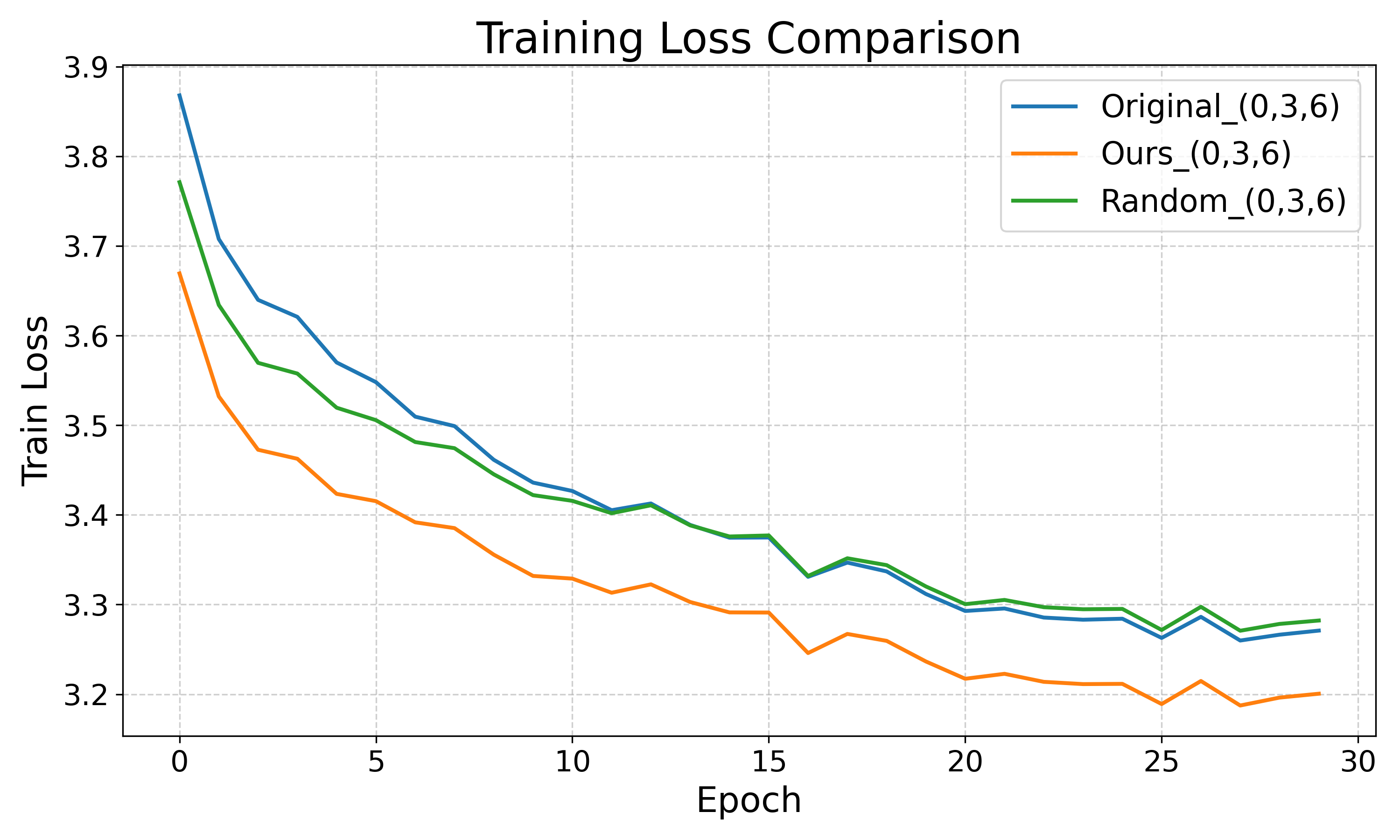}
        \caption{Pruning is applied at layers (0, 3, 6).}
        \label{fig:evit_training_loss_0_3_6}
    \end{subfigure}
    \hfill
    \begin{subfigure}{0.48\textwidth}
        \includegraphics[width=\linewidth]{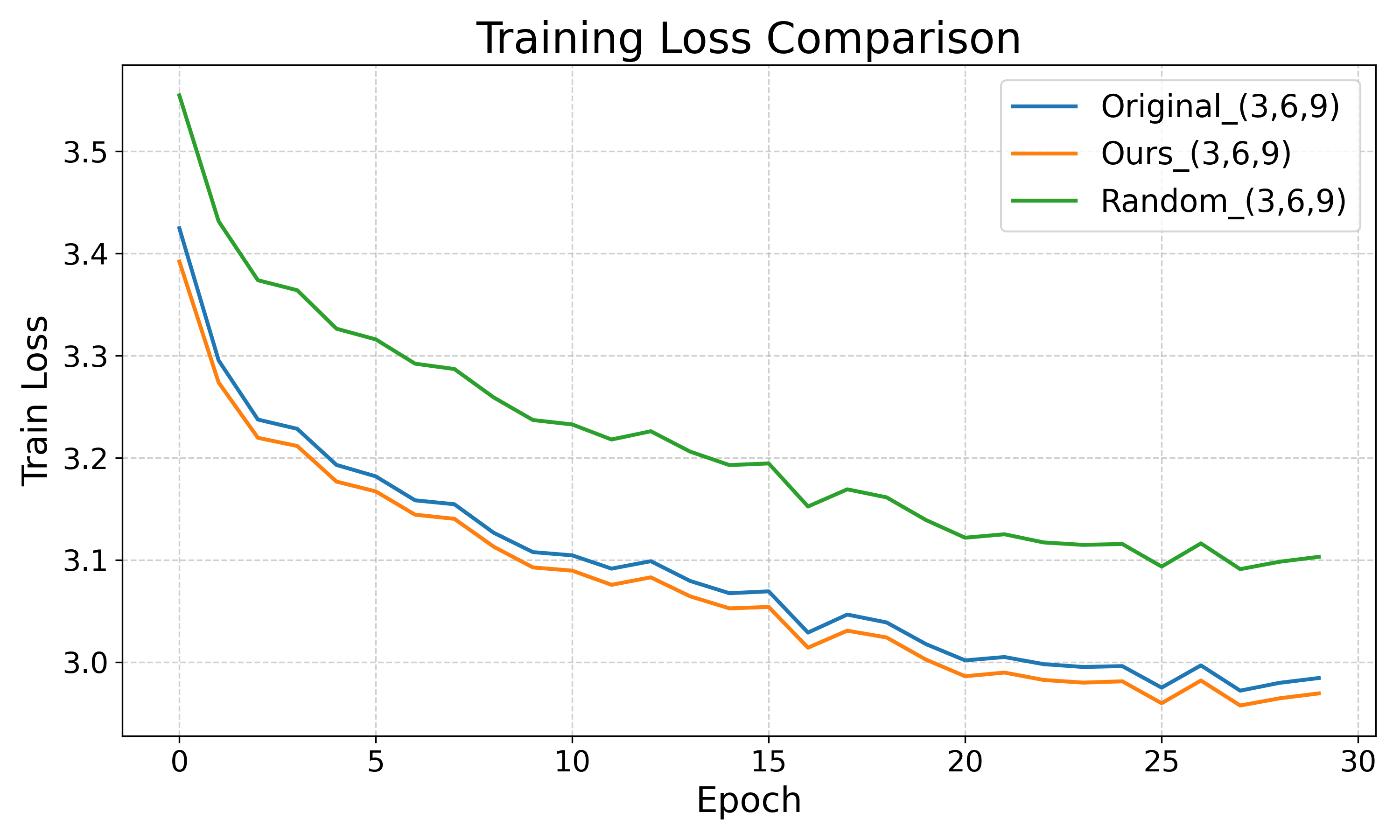}
        \caption{Pruning is applied at layers (3, 6, 9).}
        \label{fig:evit_training_loss_3_6_9}
    \end{subfigure}
    \caption{Training Loss Comparison on the EViT Framework.}
    \label{fig:evit_training_loss}
\end{figure}

\begin{wraptable}{r}{0.45\textwidth}
    \centering
    \captionof{table}{EViT fine-tuning result with a 0.7 keep rate. Our method significantly outperforms the \texttt{[CLS]} metric in early-layer pruning (0, 3, 6).}
    \label{tab:evit_acc}
    \small
    \resizebox{\linewidth}{!}{
        \begin{tabular}{c|c|ccc|c}
          \toprule
          Model & Pruning Layers & \texttt{[CLS]}  & \textbf{Ours} & Random & $GFLOPs$ \\
          \midrule
          \multirow{2}{*}{ViT-S} & 0, 3, 6 & 77.2 & \textbf{79.3} & 77.3 & 2.3 \\
          & 3, 6, 9 & 81.0 & \textbf{81.1} & 79.4 & 3.0 \\
          \bottomrule
        \end{tabular}
    }
\end{wraptable}

As shown in Table~\ref{tab:evit_acc}, our method demonstrates a significant advantage in early-layer pruning. Specifically, for the $(0, 3, 6)$ schedule, the \texttt{[CLS]} baseline achieves only 77.2\% Top-1 accuracy, which is identical to the performance of Random pruning. In contrast, our method achieves \textbf{79.3\%} accuracy, outperforming the baseline by \textbf{+2.1\%.} This demonstrates that our proposed metric achieves superior performance not only in training-free settings but also in fine-tuning scenarios.

The training loss trajectories in Fig.~\ref{fig:evit_training_loss} further clarify this phenomenon. Notably, in the (0, 3, 6) pruning scenario, the \texttt{[CLS]} metric (blue) aligns more closely with random selection (green), with its initial loss even exceeding the random baseline. In contrast, our method (orange) consistently starts with a significantly lower initial loss compared to the \texttt{[CLS]} metric. This indicates that our metric preserves more critical information from the start, avoiding the noisy signals of \texttt{[CLS]} in early layers and resulting in more effective model training. While the gap narrows for the $(3, 6, 9)$ schedule, our method remains highly competitive.

\section{Conclusion}
In this paper, we have introduced a novel, training-free token pruning metric derived from Rényi entropy, called Col-Ln. This metric can address the limitations of traditional \texttt{[CLS]}-based pruning methods, particularly their unreliability in the early layers of Vision Transformers. Through extensive experiments, we have demonstrated that our method enables more accurate identification of informative tokens from the very first layer. 

\bibliographystyle{plain} 
\small 
\bibliography{references}    


\appendix


\section{Overview}
In this supplementary material, we provide additional technical details and experimental results that were omitted from the main paper due to space constraints. Specifically, this document includes:
\begin{enumerate}
    \item \textbf{Extended visualizations:} We provide additional layer-wise attention \\
    heatmaps to further illustrate the limitations of [CLS] attention and the stability of our Col-Ln metric. (Sec.~\ref{sec:extended_vis})
    \item \textbf{Hyperparameter sensitivity analysis:} We investigate how the choice of the norm order $n$ in Rényi Entropy affects noise suppression and token importance estimation (Sec.~\ref{sec:norm_order})
    \item \textbf{Evaluation on DeiT backbones:} We present comprehensive training-free pruning results on DeiT-Small and DeiT-Base from the EViT framework to demonstrate the effectiveness of our method under the original settings proposed in the EViT paper (Sec.~\ref{sec:deit_results}).
    \item \textbf{Implementation details for fine-tuning:} We provide the detailed hyperparameters and experimental configurations used for the fine-tuning experiments on the EViT framework to ensure reproducibility (Sec.~\ref{sec:finetune_params}).
\end{enumerate}

\noindent The implementation code is available at \href{https://github.com/Wayne0758/SparseAttention}{here}.

\section{Extended Visualizations}
\label{sec:extended_vis}
In this section, we provide the attention heatmaps corresponding to Fig. 1 of the main paper to further illustrate the internal mechanisms of our metric. These heatmaps represent the importance scores assigned to each token before the pruning process.

\begin{figure}[t]
  \centering
  \includegraphics[width=\linewidth]{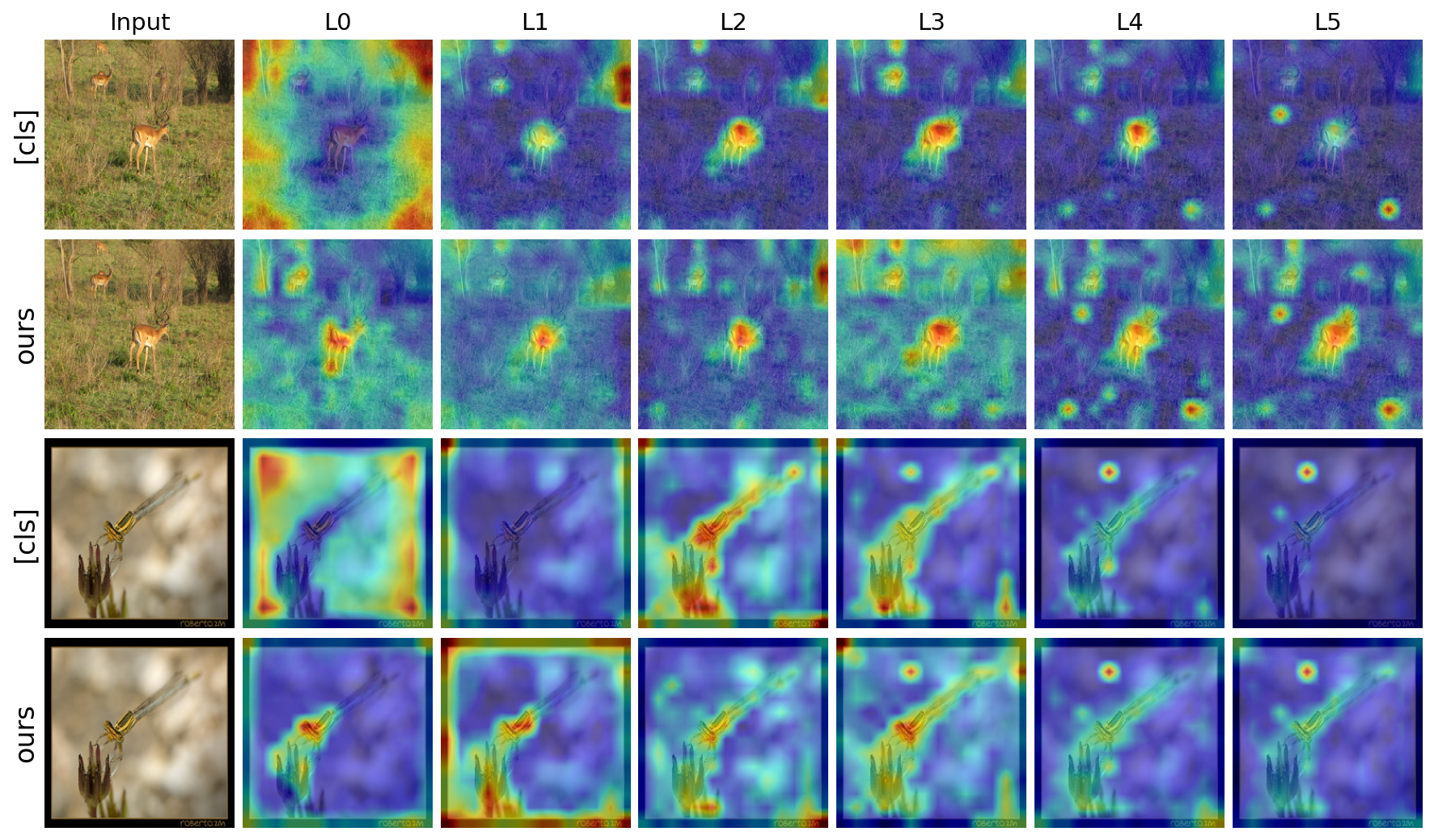}
  \caption{\textbf{Heatmap-based Attention Visualization.} This figure provides the attention heatmaps for the samples shown in Fig. 1 of the main paper. We compare the standard \texttt{[CLS]} attention (top rows) with our proposed \textbf{Col-Ln} metric (bottom rows) across the initial six layers ($L_0$--$L_5$) of a ViT-Base model. Red colors indicate higher importance scores.}
  \label{fig:supp_heatmap}
\end{figure}

As illustrated in Fig.~\ref{fig:supp_heatmap}, our method precisely highlights the primary subject (e.g., the deer and dragonfly) from the very first layer. In contrast, the \texttt{[CLS]} attention maps frequently focus on background regions rather than the salient subject and are highly susceptible to prominent edge artifacts in the early stages. This visual evidence confirms that our collective consensus mechanism provides a much more reliable and noise-resistant signal for early-layer token pruning.

\section{Impact of the Norm Order $n$}
\label{sec:norm_order}
In this section, we investigate how the choice of the norm order $n$ affects the quality of token importance estimation. As illustrated in Fig.~\ref{fig:ln-norm}, the layer-wise visualizations reveal several key patterns that justify our approach:

\begin{figure}[t]
  \centering
  \includegraphics[width=\linewidth]{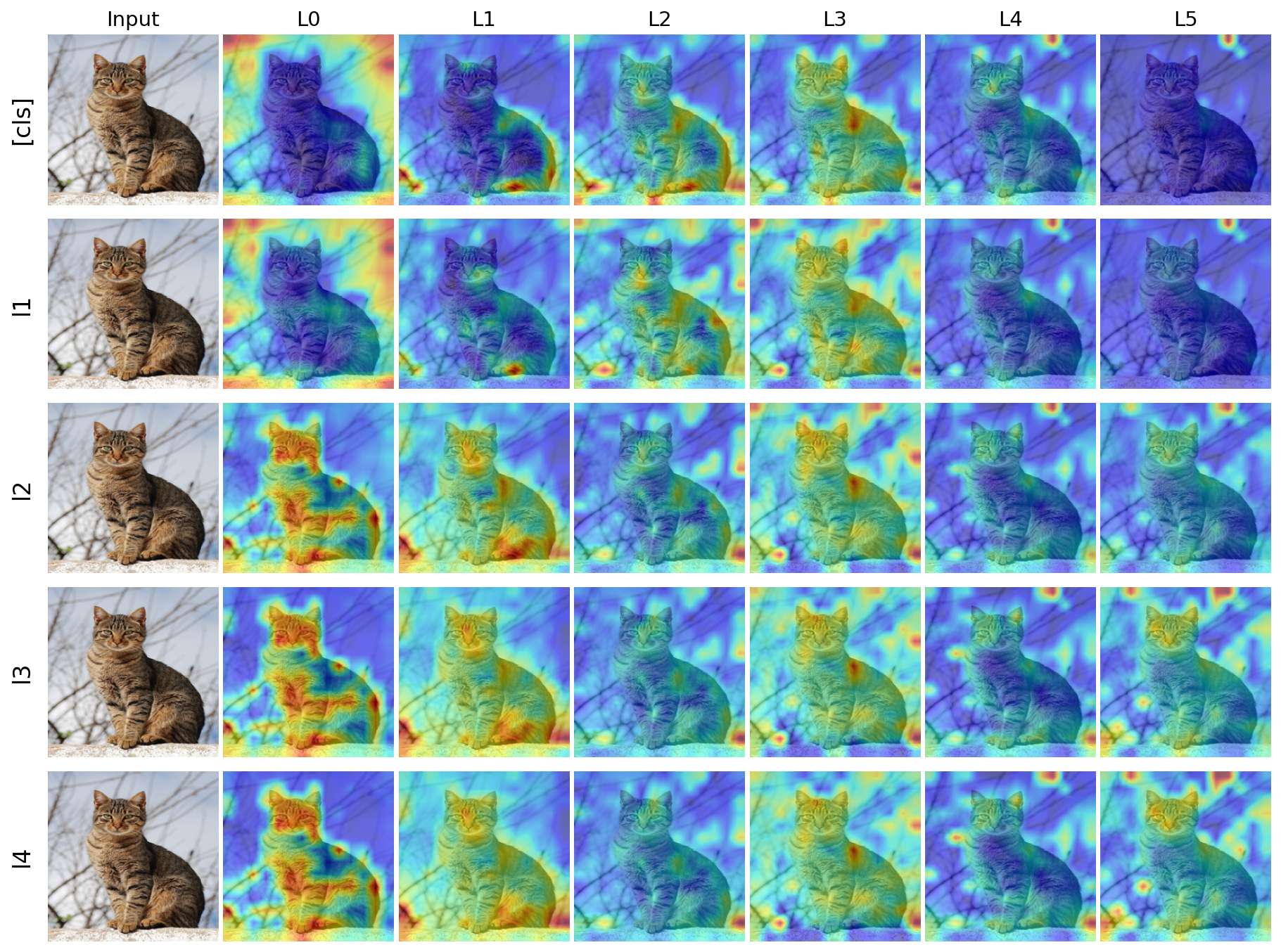}
  \caption{\textbf{Layer-wise attention visualization on ViT-Base with different norm} The top row displays traditional [CLS] attention, which is messy and noisy in early layers (L0–L3). In contrast, the bottom rows show our column-wise $\ell_n$-norms. As the norm order $n$ increases (from $\ell_1$ to $\ell_4$), the heatmaps become progressively sharper and more concentrated on the salient object.}
  \label{fig:ln-norm}
\end{figure}

\begin{itemize}
    \item \textbf{Inadequacy of the \texttt{[CLS]} Proxy:} The top row demonstrates that \texttt{[CLS]} attention is highly messy and noisy in the early layers ($L_0$--$L_3$). This visual evidence confirms our hypothesis regarding \textit{semantic immaturity}, where the \texttt{[CLS]} token fails to provide a reliable pruning signal before it has successfully aggregated global context.
    
    \item \textbf{Effectiveness of Column-wise Norms:} In contrast, the column-wise $\ell_n$-norms identify the target object starting from the very first layer. This indicates that collective consensus among patches is a more robust indicator of importance than the state of the \texttt{[CLS]} token in shallow layers.
    
    \item \textbf{Noise Suppression via Higher-Order Norms:} The evolution from $\ell_1$ to $\ell_4$ illustrates a clear refinement in focus. As visualized in Fig.~\ref{fig:ln-norm}, the heatmap remains relatively blurry and messy at $\ell_1$; however, as the norm order $n$ increases, the heatmaps become increasingly clearer and more sharply concentrated on the semantic core of the object. This shows that higher-order norms effectively suppress these distractions to highlight the most discriminative features.
\end{itemize}

Based on these observations, we recommend utilizing norm orders in the range of \texttt{$n \in [2, 4]$} for our Col-Ln metric. Within this range, the indicator maintains a high signal-to-noise ratio, providing precise and noise-resistant guidance for token selection across the entire network.

\section{EViT Inference Results on DeiT}
\label{sec:deit_results}

While we utilized ViT-Small and ViT-Base in the main paper to maintain consistency across different experiments, in this section, we provide additional inference results using \textbf{DeiT-Small} and \textbf{DeiT-Base}. These models were the original architectures evaluated in the EViT paper. As shown in Table~\ref{tab:deit_small_results} and Table~\ref{tab:deit_base_results}, our Col-Ln metric consistently outperforms the [CLS] baseline under the original experimental settings of EViT, further demonstrating the robustness of our collective consensus mechanism in a training-free scenario.

\begin{table}[t]
\centering
\footnotesize
\setlength{\tabcolsep}{4pt}
\caption{EViT inference on \textbf{DeiT-Small}.}
\label{tab:deit_small_results}
\begin{tabular}{c|c|ccc|c}
\toprule
Keep Rate ($r$) & Pruning Layers & \texttt{[CLS]} & \textbf{Ours} & Random & $GFLOPs$ \\
\midrule
DeiT-Small  & -- & 79.8 & 79.8 & 79.8 & 4.6 \\
\midrule
\multirow{2}{*}{0.7} & 3, 6, 9 & \textbf{78.5} & \textbf{78.5} & 77.5 & 3.0 \\
 & \textbf{0, 3, 6} & 64.7 & \textbf{75.1} & 74.5 & 2.3 \\
 \midrule
\multirow{2}{*}{0.8} & 3, 6, 9 & \textbf{79.3} & \textbf{79.3} & 78.5 & 3.5 \\
 & \textbf{0, 3, 6} & 72.1 & \textbf{77.7} & 77.3 & 2.9 \\
\midrule
\multirow{2}{*}{0.9} & 3, 6, 9 & \textbf{79.7} & \textbf{79.7} & 79.3 & 4.0 \\
 & \textbf{0, 3, 6} & 77.4 & \textbf{79.1} & 78.9 & 3.7 \\
\bottomrule
\end{tabular}
\end{table}

\begin{table}[t]
\centering
\footnotesize
\setlength{\tabcolsep}{4pt}
\caption{EViT inference on \textbf{DeiT-Base}.}
\label{tab:deit_base_results}
\begin{tabular}{c|c|ccc|c}
\toprule
Keep Rate ($r$) & Pruning Layers & \texttt{[CLS]} & \textbf{Ours} & Random & $GFLOPs$ \\
\midrule
DeiT-Base  & -- & 82.0 & 82.0 & 82.0 & 4.6 \\
\midrule
\multirow{2}{*}{0.7} & 3, 6, 9 & \textbf{80.6} & \textbf{80.6} & 78.8 & 3.0 \\
 & \textbf{0, 3, 6} & 71.3 & \textbf{76.6} & 75.7 & 2.3 \\
 \midrule
\multirow{2}{*}{0.8} & 3, 6, 9 & \textbf{81.3} & \textbf{81.3} & 80.3 & 3.5 \\
 & \textbf{0, 3, 6} & 77.0 & \textbf{79.5} & 79.0 & 2.9 \\
\midrule
\multirow{2}{*}{0.9} & 3, 6, 9 & 81.7 & \textbf{81.8} & 81.2 & 4.0 \\
 & \textbf{0, 3, 6} & 80.2 & \textbf{81.1} & 80.8 & 3.7 \\
\bottomrule
\end{tabular}
\end{table}

\section{Fine-tuning Implementation Details}
\label{sec:finetune_params}

To ensure the reproducibility of our results, we provide the detailed hyperparameter configuration used for the fine-tuning experiments on the EViT framework. The model was fine-tuned for 30 epochs on the ImageNet-1K dataset. The specific parameters used in our training script are summarized in Table~\ref{tab:hyperparams}.

\begin{table}[h]
\centering
\caption{\textbf{Hyperparameters for EViT Fine-tuning.} These settings were applied to the ViT-Small (augreg) backbone used in our comparative analysis.}
\label{tab:hyperparams}
\begin{tabular}{lc}
\toprule
\textbf{Hyperparameter} & \textbf{Value} \\ \midrule
Backbone Model & \texttt{vit\_small\_patch16\_augreg} \\
Input Resolution & $224 \times 224$ \\
Total Batch Size & 2048 \\
Optimizer & AdamW \\
Learning Rate (Base) & $2 \times 10^{-5}$ \\
Min Learning Rate & $2 \times 10^{-6}$ \\
Weight Decay & $1 \times 10^{-6}$ \\
LR Scheduler & Cosine decay \\
Training Epochs & 30  \\
Warmup Epochs & 0 \\
Pruning Start Epoch & 0 \\
Keep Rate ($r$) & 0.7  \\
Rescue Ratio ($c$) & 0.8 \\
\bottomrule
\end{tabular}
\end{table}

\end{document}